# An Image Based Visual Servo Approach with Deep Learning for Robotic Manipulation

Jingshu Liu, Yuan Li

Beijing Institute of Technology, Beijing, China
E-mail: liujingshu2014@163.com;
liyuan@bit.edu.cn

**Abstract:** Aiming at the difficulty of extracting image features and estimating the Jacobian matrix in image based visual servo, this paper proposes an image based visual servo approach with deep learning. With the powerful learning capabilities of convolutional neural networks(CNN), autonomous learning to extract features from images and fitting the nonlinear relationships from image space to task space is achieved, which can greatly facilitate the image based visual servo procedure. Based on the above ideas a two-stream network based on convolutional neural network is designed and the corresponding control scheme is proposed to realize the four degrees of freedom visual servo of the robot manipulator. Collecting images of observed target under different pose parameters of the manipulator as training samples for CNN, the trained network can be used to estimate the nonlinear relationship from 2D image space to 3D Cartesian space. The two-stream network takes the current image and the desirable image as inputs and makes them equal to guide the manipulator to the desirable pose. The effectiveness of the approach is verified with experimental results.

## 1. INTRODUCTION

With the development of robotic technologies, the application of robots in manufacture and daily life is more and more extensive, and robot control has become a hot field in recent scientific research. To achieve intelligent control of robots, we need to give them the ability to sense the environment. Vision sensors are an important source of information for robots and provide a large amount of external information. It is feasible to control robots by using abundant visual information from the external environment. Visual servo control uses visual information to control the position and orientation of the robot. The goal is to control the robot to reach the desirable pose quickly and precisely.

Regarding the structure of the control system, visual servo approaches can be divided into image based visual servo (IBVS) and position based visual servo (PBVS) [1]. The image based visual servo forms a closed loop control system in the 2-D image space, extracting image features from visual information of the observed object in the image space, and then calculate the error compared with the desirable image features. Then the control law is designed according to the error. The position based visual servo forms a closed loop control system in the 3-D Cartesian space, calculating the position and orientation of the observed object from visual information in three-dimensional space, and compare the error between the expected position and orientation value, then the control law is designed according to the error [2][3].

This paper engages the feature extraction issue in image based visual servo. IBVS forms a closed loop in the 2-D image space, which is insensitive to the camera's calibration error and robot model error [4]. It has high steady state control accuracy and is often applied in various kinds of "eye-in-hand" robot control systems. This kind of control system have received widespread attention from researchers in the field of robotic servo because of its low cost, easy ease of use and maintenance [5][6][7].

Traditional IBVS system is based on image features extracting from images of the observation target, and then the interaction matrix (also called as feature Jacobian) which reflects the mapping from image space to task space is constructed to generate the corresponding control law. In this procedure, the selection and extraction of image features is a challenged problem. Sometimes the field of view does not consist of sufficient image features of the observation target so that we need to introduce additional visual features actively (e.g. structured light [8][9]). Some researchers proposed automatic approaches to extract features from images, however, these approaches have strict restrictions and can just be applied in specific situations [10][11].

It is a challenge to construct Jacobian matrix, too. The derivation process of Jacobian is cumbersome and complicated due to its nonlinearity. What's more, real-time estimation of Jacobian is needed in most actual control tasks because of its time variation, which leads to a large amount of calculation. To solve this problem, some researchers have proposed to introduce the neural network into the visual servo control [12]. The neural network can integrate visual information and generate control law directly, which frees us from the analysis and estimation of the feature Jacobian [13][14]. However, this approach still requires extracting features from the image manually as training samples in advance.

Aiming at the above problems, we propose a robotic visual servo method based on image and deep learning, which realizes the control of 4 degrees of freedom of the robot manipulator (3 of translation and 1 of rotation), and we verifies the effectiveness and accuracy of this method in experiments. A two-stream network based on convolutional neural network is proposed to extract the image features of the observed targets in the current situation (position and orientation) and compare them with the features of the ideal situation. Then the control law is determined to make the manipulator move from the existing pose to the ideal position quickly and accurately. With this two-stream network we do not need to manually select image information or derive the mapping relationship from image space to task space, which greatly facilitates the visual control design.

The remainder of this paper is organized as follows. In Section 2, the framework of IBVS and deep learning is introduced. Then in Section 3, the visual control task is defined and our network architecture as well as control scheme are given. After that in Section 4, experiments were conducted to verify our control scheme. Finally, conclusions are presented in Section 5.

## 2. FRAMEWORK OF IBVS AND DEEP LEARNING

### 2.1. IBVS

The image based visual servo defines the task error in 2-D image space therefore the velocity of robot's end effector (manipulator) can be regulated using visual feedback. In visual servo system, considering an "eye-in-hand" system, in which a camera is fixed onto robot's end effector and move together, the state-space model can be described as:

$$\begin{cases} v(t) = \dot{r}(t) \\ s(t) = f(r) \end{cases} \quad (1)$$

where $r(t)$ is the camera pose; $v(t)$ is the camera velocity (also velocity of the manipulator); $s(t)$ is the image features and $f$ represents the imaging relationship from 3-D to 2-D space. The relationship between camera velocity and image features can be obtained by calculating time derivative of (1):

$$\dot{s} = L_s v \quad (2)$$

in which $L_s = \dfrac{\partial f(r)}{\partial s}$ is the feature Jacobian.

The structure of IBVS system is shown in Fig. 1. In Fig. 1, $s$ is current visual features, $s^*$ is its value at desirable position and rotation; error $e$ is defined by:

$$e = s - s^* \quad (3)$$

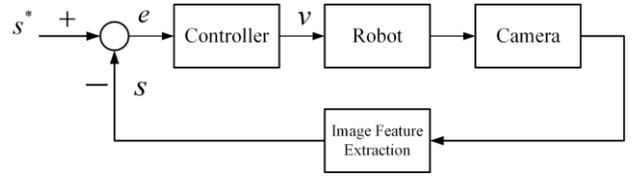

**Fig. 1** Image based visual servo scheme.

The aim of all IBVS control schemes is to minimize the error between current visual features and desirable features. Considering the time derivative of (3) we can get $\dot{e} = \dot{s}$ and apply it to (2):

$$\dot{e} = L_s v \quad (4)$$

Imposing an exponential decrease of the error (i.e. $\dot{e} = -\lambda e$), we obtain using (4):

$$v = -\lambda L_s^+ e \quad (5)$$

where $\lambda$ is a positive gain and $L_s^+$ is the pseudo-inverse matrix of $L_s$.

Equation (5) describes the control law of IBVS, of which the design of feature Jacobian is a key issue. However, in actual situations, the matrix $L_s^+$ is often difficult to calculate directly and needs real-time estimation.

### 2.2. Deep Learning

Deep learning is now a hotspot and has achieved breakthrough achievements in many fields. Its basic idea is to use supervised or unsupervised strategies to learn multi-level representations and features in hierarchical structures for classification or regression. Convolutional neural networks (CNN) are the most extensive model for solving computer vision problems with deep learning methods. Since LeCun et al. first introduced CNN to the field of image identification in 1989 [15] and proposed LeNet-5 in 1998 [16], CNN has been widely used in various tasks in the field of computer vision.

Compared to traditional shallow machine learning techniques, deep web can learn complex features from a large number of samples and combine low-level features to obtain features that are more advanced. Taking CNN as an example, CNN can extract and combine image features through convolutional and pooled layers. At the same time, it has similar excitation functions and fully connected layer structures that can fit the complex nonlinear relationship as traditional neural networks.

CNN can both autonomously extract image features from samples and fit the nonlinear mapping from 2D image space to 3D task space, so there is great potential for CNN in IBVS. Generally, supervised learning can be divided into two categories: classification and regression, and CNN are very commonly seen in the former such as face recognition. In recent years, researchers have begun to use the regression function of CNN in the field of robotics. Lenz et al. [17] present a two-step cascaded system with two deep networks for detecting robotics grasps. Redmon and Angelova simplify the network structure in [17] and propose a one-step regression approach to graspable bounding based on CNN [18].

Inspired by this, we proposes a two-stream network based on CNN to realize the visual control of robots. The CNN in our network performs one-step regression from image to 3-D space parameters.

## 3. VISUAL SERVO DEVELOPMENT

### 3.1. Definition of Task

In robotic application, the task of pick and place is the most fundamental for the manipulator. Usually, this task can be divided into the following three subtasks: 1) guiding the robot manipulator to the desirable position and orientation; 2) letting it grab the target; 3) moving the manipulator to another place. Our paper will focus on the first subtask.

In this paper, the vision system is of the "eye in hand" type where a camera is fixed onto the manipulator. The visual system configuration is shown in Fig.2. The camera is mounted on the end effector of robot so that the camera frame is parallel to the end effector frame with some offset. The observed target is placed on a platform and the target frame is parallel to the robot base frame. The camera captures images of the observation target in real time, and in the control network features are extracted from the images to generate control law for the position and orientation of the manipulator. Our system realizes the control of 4 degrees of freedom of the manipulator: the translation along x, y, z-axis and rotation of z-axis corresponding to robot base frame.

### 3.2. Network Structure

We propose a two-stream control network based on CNN, as shown in Fig.3. There are two inputs in the two-stream network: one is the image of target when the manipulator is set in ideal pose (Image1), the other is the image in current pose (Image2), and the output is the control law for the four degrees of freedom respectively of the robot. Correspondingly, there are two same CNNs, which are trained offline in advance.

The architecture of CNN in Fig.3 is shown in Fig.4. We derive our CNN architecture from a version of the widely adopted convolutional network proposed by Krizhevsky et al. [18][19]. Our architecture has five

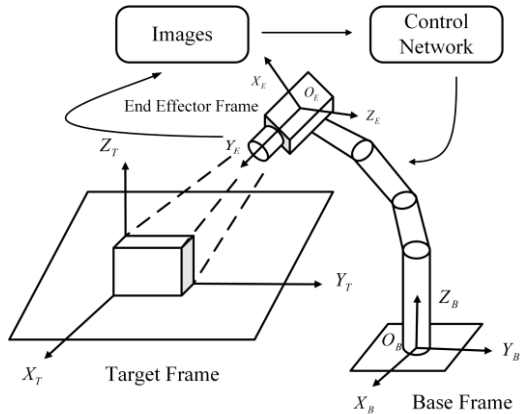

**Fig. 2** Visual system configuration.

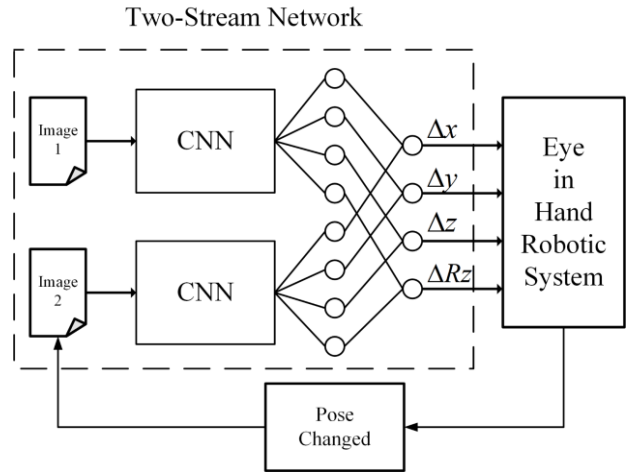

**Fig. 3** Two-stream control system.

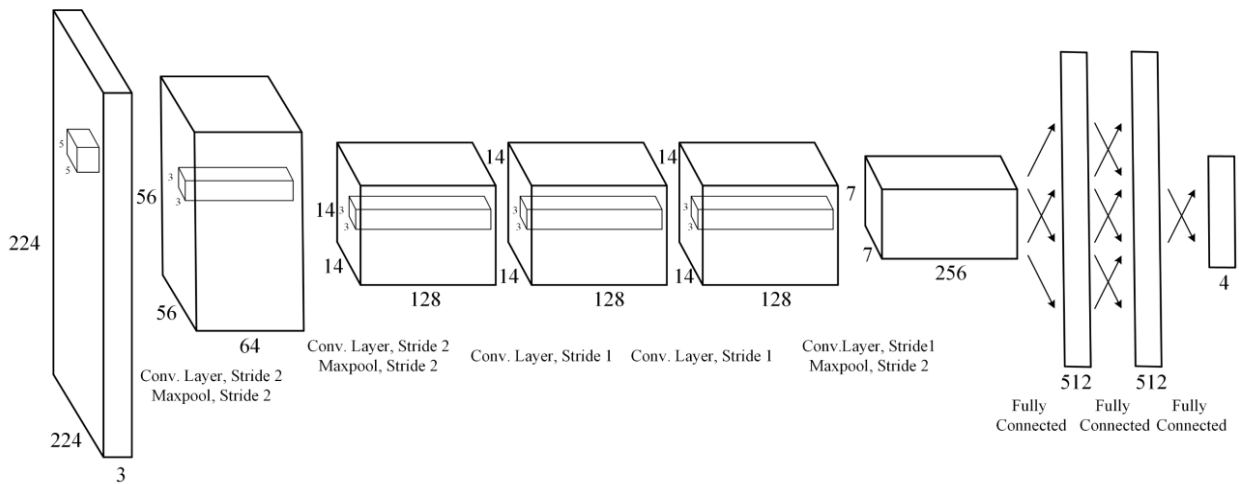

**Fig. 4** CNN architecture.

An Image Based Visual Servo Approach with Deep Learning for Robotic Manipulation

convolutional layers followed by three fully connected layers and the convolutional layers are interspersed with activation function and max pooling layers at various stages. The architecture takes an image and then regression is done to figure out the four parameters of position and orientation corresponding to the image.

### 3.3. Control Scheme

In our control system in Fig.3, one CNN takes the desirable image as input and the other takes the current image, and the outputs of these two CNNs are pose parameters at corresponding position and orientation, that is the four pose parameters under desirable pose (denoted as $r^*$) and current pose (denoted as $r_c$), with

$$r^* = (x^*, y^*, z^*, Rz^*) \quad (6)$$

$$r_c = (x_c, y_c, z_c, Rz_c) \quad (7)$$

Here, $x$, $y$ and $z$ in $r$ means the offset of the robot end effector frame to base frame and $Rz$ means its rotation around z-axis of base frame.

Our paper focuses on how to move the robot to the designated location. So we don't pay attention to the absolute value of the above parameters. What we care about is the error between $r^*$ and $r_c$. A connected layer is constructed to calculate the error between the corresponding DOF of the two network's output. The error is defined as:

$$e = r_c - r^* = (x_c - x^*, y_c - y^*, z_c - z^*, Rz_c - Rz^*) \quad (8)$$

And we choose a simple proportional control scheme $v = (\Delta x, \Delta y, \Delta z, \Delta Rz) = -\lambda e$. That is:

$$\Delta x = -\lambda_1 (x_c - x^*) \quad (9)$$

$$\Delta y = -\lambda_2 (y_c - y^*) \quad (10)$$

$$\Delta z = -\lambda_3 (z_c - z^*) \quad (11)$$

$$\Delta Rz = -\lambda_4 (Rz_c - Rz^*) \quad (12)$$

Choosing probable proportional gain, each DOF is calculated and controlled separately so that decoupled control is achieved. In addition, there is no need for camera calibration because what we only concern is image features.

### 4. EXPERIMENTS AND ANALYSIS

Experiment is done in order to verify the effectiveness of the visual servo approach we proposed. The experimental platform is shown in Fig.5. The robot in our experiments is YASKAWA NOTOMAN MH5 which is an industrial 6 DOF manipulator. The actual value of six degrees of freedom of the manipulator relative to the robot base frame can be obtained from the control panel. These data can be used in network training. The camera is attached to the manipulator. The observation target is placed on the platform. It is an irregularly shaped metal block, whose color is similar to the color of the background platform. It is difficult to extract the features of the irregular shape under such conditions by using traditional image processing method, which can reflect the superiority of deep learning that extracts features autonomously.

### 4.1. Network Training

The network that needs to be trained in advance is the CNN in the control system of Fig.3, and its structure is shown in Fig. 4. Move the manipulator to different position and orientation, take photos of the target and record the DOF parameters on the control panel as training samples. In our experiment, the target frame is parallel to the robot base frame. The four degrees of freedom controlled in experiment are the three translation of x, y, z-axis and rotation around the z-axis of the robot base frame.

Due to the limitations of the experimental conditions, we gather 400 images as training samples. Part of the samples are shown in Fig.6. The original image size captured by the camera is 480×360, and is resized to 224×224 in training to reduce the amount of network parameters. We choose the mean square error as the loss function and use the stochastic gradient descent method to train. The learning rate is 0.0005 across all the layers and with a weight decay of 0.001. Between the fully connected layers, dropout is adopted as a kind of regularization. The probability of dropout is 0.5. We train our model for 100 epochs with the batch size of 20.

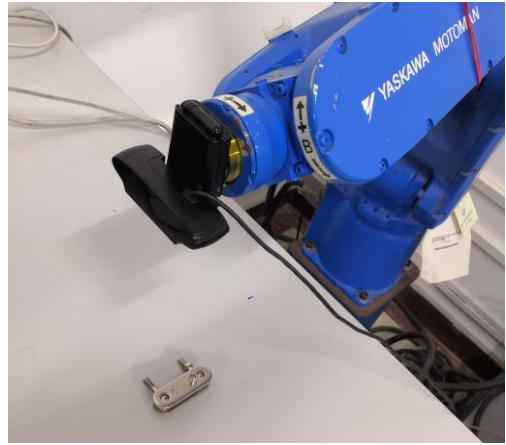

**Fig. 5** Experiment platform: YSKAWA MOTOMAN.

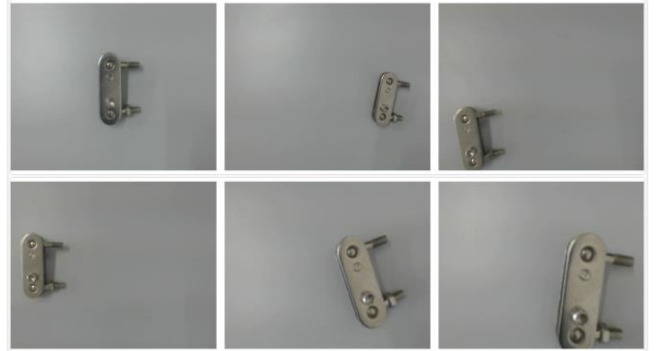

**Fig.6** Some training samples.

After disrupting the order of 400 image samples, we randomly selected 20 samples as test set and the rest as training set. After the network training is completed, the result of network training is examined on the test set. Comparing the pose parameters of the network output with the actual value of test set, the results are as follows: for translation, the average absolute error in the x direction is 3.72mm, 3.58mm in y direction and 4.02mm in z direction; for rotation around z-axis the average absolute error is 3.02 degree. The length of the target is about 55mm and its width is about 33mm. The range of distance from camera to the target is about 250-400mm. This indicates that error is within acceptable range and the network after training is of high accuracy.

### 4.2. Visual Servo Experiment

After training, the network is utilized in the visual control of the robot manipulator. The configuration of the vision system is the same as that in the network training as shown in Fig.5. The placement of the observed target and the way the camera mounted are the same.

In our visual control experiment, the target is placed in a fixed position. Choose a desirable pose and capture the image of the target in that pose as one input to the two-stream control network. Moving the manipulator to arbitrary pose, the image observed in the current pose is captured as the other input to the control network. As shown in Fig.7, under visual control, the manipulator of robot can move from the initial pose to the desirable pose, and the image features change from the initial state to the desirable state.

In the experiment, the proportional control gain $\lambda$ is selected as 0.03. Record the changing process of the pose parameters in the control procedure and draw the graphs. The results are shown in Fig.8-11.

Fig. 8-10 shows the error curve of the three translation directions, and Fig.11 shows the error curve of rotation angle around z-axis of the robot base frame. As can be seen from these figures, the error of each parameter is reduced to about zero in less than 15 steps and remain stable. To ensure accuracy, we take some other different poses as initial and desirable states and all obtain good results. This indicates that the approach we proposed can control the manipulator to move to the desirable position and orientation at a relatively high speed.

There exists error in both CNN training and visual servo experiment. The most probable reason is that the amount of training samples is relatively small due to the limitations of the experimental conditions. Enlarging the amount of training samples will make the training result of CNN more precise. Meanwhile, improving the illumination condition can make image features more conspicuous and may also do some help to experimental result.

### 5. CONCLUSIONS

Image based visual verso is widely applied in robot control in industrial occasions. In traditional image based visual servo approaches, the selection and extraction of image features and the construction of nonlinear mapping from image space to task space (image

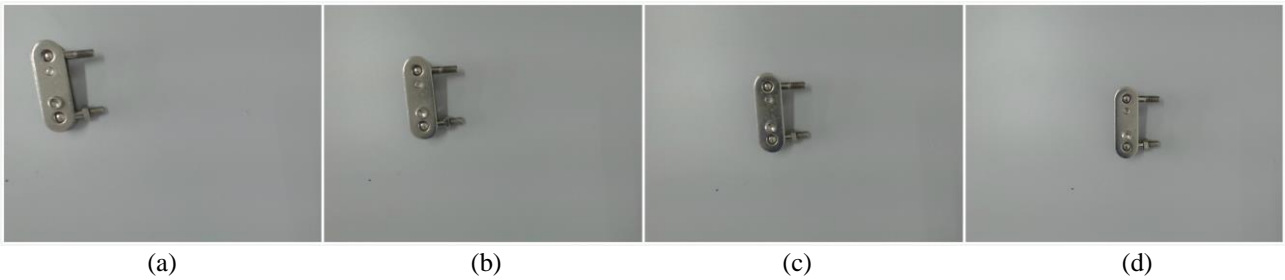

(a)          (b)          (c)          (d)

**Fig. 7** Image of some steps in visual servo: (a) initial state; (b) and (c) intermediate steps; (d) final state (desirable state).

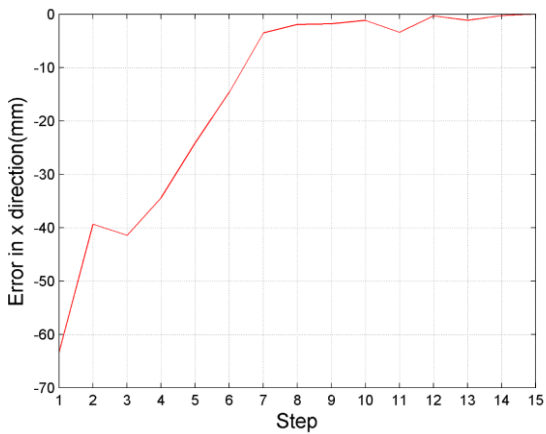

**Fig. 8** Error in translation along x-axis.

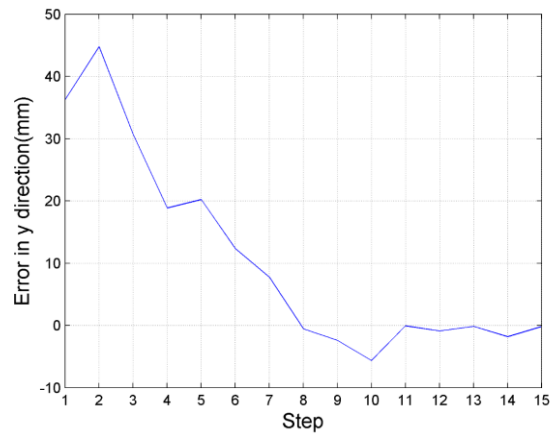

**Fig. 9** Error in translation along y-axis.

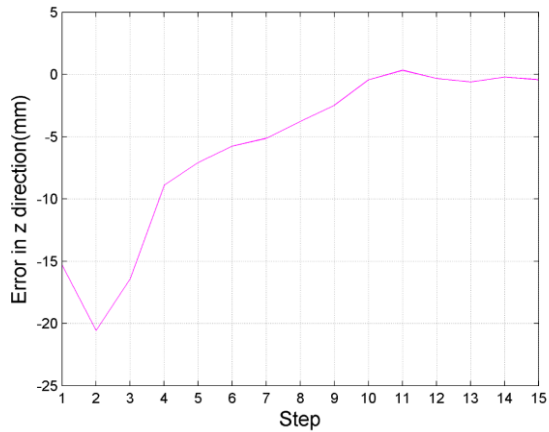

**Fig. 10** Error in translation along z-axis.

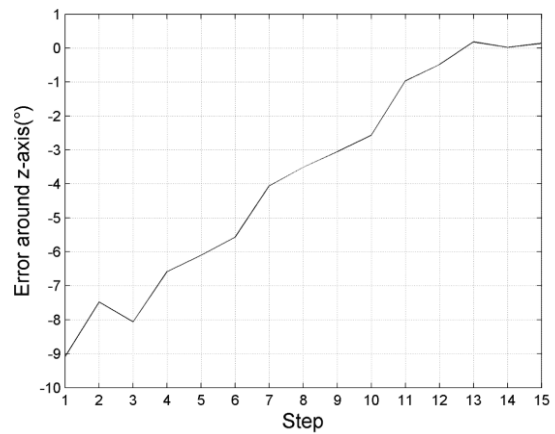

**Fig. 11** Error in rotation around z-axis.

Jacobian) are two key points. As the complexity of the environment and tasks increases, the difficulty of these two issues will enhance greatly. The convolutional neural network can extract the key information in images well and can fit various nonlinear mapping relationships. Therefore, introducing the convolutional neural network into the visual control can greatly simplify and solve the above problems. In this paper, an IBVS approach based on deep learning is proposed. The convolutional neural network is used to extract features from the image to realize direct mapping from image to 3-D pose parameters.

In this paper, we propose an image based visual servo approach with deep learning for robotic manipulation. We design a two-stream network based on convolutional neural network and the visual control scheme. The convolutional neural network is used to calculate the pose parameter values, and then a connected layer is used to calculate the error of each parameter and generating a corresponding control law which enables independent decoupled control of each degree of freedom. We gather sample images and build a database for network training. After training the network is tested in experiment. The experimental results show that the control method has a good effect.

ACKNOWLEDGEMENTS

This work is supported by the National Science Foundation of China under Grants 61472037, 61433003.